\pdfoutput=1

\documentclass[11pt]{article}

\usepackage[preprint]{acl}

\usepackage{times}
\usepackage{latexsym}
\usepackage[capitalize]{cleveref}
\usepackage{enumitem}

\newcommand{\intern}{InternLM2-Math-20B}
\newcommand{\llama}{Llama2-Chat-70B}
\newcommand{\mixtral}{Mixtral-8$\times$7B}
\newcommand{\chatgpt}{GPT-3.5-Turbo}
\newcommand{\gptf}{GPT-4}
\newcommand{\gemini}{Gemini Pro}

\usepackage[T1]{fontenc}

\usepackage[utf8]{inputenc}

\usepackage{microtype}

\usepackage{inconsolata}

\usepackage{graphicx}
\usepackage{subfigure}
\usepackage{soul}
\usepackage{booktabs}
\usepackage{colortbl}
\usepackage{multirow}
\usepackage{multicol}
\usepackage{makecell}
\usepackage{tabularx}

\title{\texttt{CLEAR}: Can Language Models Really Understand Causal Graphs?}

\author{
 \textbf{Sirui Chen\footnotemark[1]\textsuperscript{1}},
 \textbf{Mengying Xu\textsuperscript{2}},
 \textbf{Kun Wang\textsuperscript{2}},
\\
 \textbf{Xingyu Zeng\textsuperscript{2}},
 \textbf{Rui Zhao\textsuperscript{2}},
 \textbf{Shengjie Zhao\textsuperscript{1}},
 \textbf{Chaochao Lu$\footnotemark[2]$\textsuperscript{3}}
\\
\\
 \textsuperscript{1}Tongji University,
 \textsuperscript{2}SenseTime Group,
 \textsuperscript{3}Shanghai AI Laboratory
 \\
 \small{
\texttt{chensirui@pjlab.org.cn},  \texttt{\{xumengying, wangkun, xyzeng\}@sensetime.com}, \texttt{luchaochao@pjlab.org.cn}
 }
}

\begin{document}
\maketitle

\renewcommand*{\thefootnote}{\fnsymbol{footnote}}
\footnotetext[1]{Work done when interning at Shanghai AI Laboratory.}
\footnotetext[2]{Corresponding author.}

\renewcommand{\thefootnote}{\arabic{footnote}}
\setcounter{footnote}{0}

\begin{abstract}
Causal reasoning is a cornerstone of how humans interpret the world. To model and reason about causality, causal graphs offer a concise yet effective solution.  Given the impressive advancements in language models, a crucial question arises: can they really understand causal graphs? To this end, we pioneer an investigation into language models' understanding of causal graphs. Specifically, we develop a framework to define causal graph understanding, by assessing language models' behaviors through four practical criteria derived from diverse disciplines (e.g., philosophy and psychology).
We then develop \texttt{CLEAR}, a novel benchmark that defines three complexity levels and encompasses 20 causal graph-based tasks across these levels. Finally, based on our framework and benchmark, we conduct extensive experiments on six leading language models and summarize five empirical findings. Our results indicate that while language models demonstrate a preliminary understanding of causal graphs, significant potential for improvement remains. Our project website is at \texttt{\url{https://github.com/OpenCausaLab/CLEAR}}.
\end{abstract}

\section{Introduction}
Causal reasoning is fundamental to how humans understand the world and solve challenges \citep{sloman2009causal}. The ability to reason causally allows us to explain phenomenon and predict the future \citep{woodward2005making,pearl2009causality,bunge2017causality}. There are various causal models used to investigate and represent causation, including mathematical equations, logical statements, and causal graphs \citep{pearl2018book}. Among them, causal graph gains widespread adoption due to its intuitive and concise representation of complex causal relationships \citep{pearl1995causal,pearl1998graphs}. 

\begin{figure}[t!]
\centering  
\includegraphics[width=.5\textwidth]{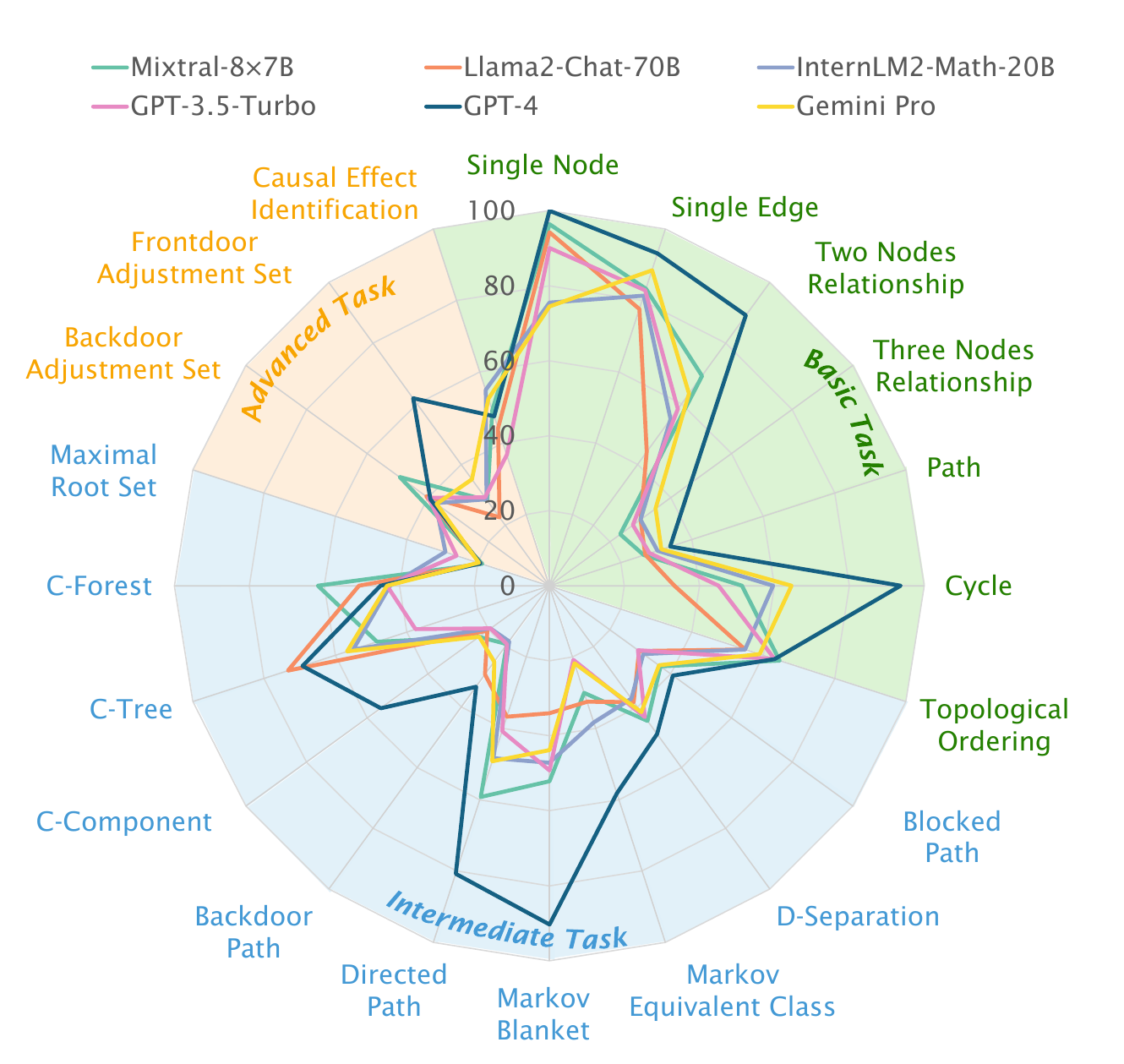}
\caption{\textbf{Performance of six leading language models across 20 diverse tasks in \texttt{CLEAR}.} Further details on the experimental results can be found in \cref{sec:experiment}.}
\label{fig:overall_radar}
\end{figure}

A causal graph is essentially a bayesian network where each node represents a variable, and the directed edges denote definite or possible causal relationships between variables \citep{helmert2004planning}. Understanding causal graphs is essential, as it enables us to grasp the relationships between variables \citep{kocaoglu2017experimental}. Furthermore, causal graphs can be leveraged for probability calculation \citep{kleinberg2013causality}, providing solutions to problems across all three rungs of the \emph{ladder of causation} (i.e., association, intervention, and counterfactuals) \citep{pearl2018book}. With the rapid advancement of language models, there has been a surge in research exploring their ability to solve graph-related problems \citep{zhang2023graph,chai2023graphllm,fatemi2023talk,ye2023natural,zhang2023llm4dyg,besta2024graph,chen2024exploring,wang2024can,luo2024graphinstruct}. In contrast to the abundant research on general graph problems, the ability of language models to understand causal graphs is yet to be investigated. Therefore, this paper aims to shed light on the question: \emph{Can language models really understand causal graphs?}

Addressing this question poses three major challenges: 
(1) What does it mean for a model to understand causal graphs?
(2) How to design a causal graph-based benchmark that can measure a model's understanding?
(3) How to quantify a model's understanding when presented with causal graphs?

In this work, we first propose a framework to evaluate language models' \emph{understanding} of causal graphs, by establishing four criteria: performance exceeding random guesses, robustness against question types, correct utilization of causal definitions, and performance constrained by task dependence. These criteria draw on insights from machine learning, philosophy, and psychology, providing a multidisciplinary approach to evaluating the comprehension of causal graphs by language models.
Next, we construct the \texttt{CLEAR}, a novel benchmark created specifically for evaluating how well language models understand causal graphs. 
Finally, guided by our proposed framework of \emph{understanding} in causal graphs, we systematically evaluate models' performance on \texttt{CLEAR} across all four criteria. To ensure a diverse evaluation, we select six leading models and utilize four prompts (e.g., in-context learning (IcL) \citep{brown2020language}). Our extensive experiments yield the following key findings:
\begin{enumerate}[itemsep=2pt,topsep=2pt,parsep=0pt]
    \item The model's ability to handle different causal graph-based tasks is uneven, exhibiting notable weaknesses in specific areas (Figure \ref{fig:overall_radar}).
    \item Language models have a preliminary understanding of causal graphs (Figure \ref{fig:overall_level_performance}), and are observed to focus on key information required to deduce the correct answer (Figure \ref{fig:counterfactual}).
    \item Model performance is sensitive to the question type. A model's understanding of causal graphs might be artificially inflated if evaluation relies on limited types (Figure \ref{fig:robustness_question_types}).
    \item Models exhibit a capacity for utilizing both explicit and implicit concepts related to causal graphs, and their proficiency with these concepts varies considerably (Figure \ref{fig:definition_IcL_overall}).
    \item The performance of most models is not constrained by task dependency (i.e., although Task B depends on Task A, performance on Task B often exceeds that on Task A), showcasing a notable divergence in their performance trends. This might suggest heterogeneity in knowledge representation and application across different models (Figure \ref{fig:task_performance_constrain}).
\end{enumerate}

Overall, we make four main contributions:
\begin{itemize}[itemsep=2pt,topsep=2pt,parsep=0pt]
    \item We make, to the best of our knowledge, the first-ever attempt to evaluate language models' capacity for understanding causal graphs.
    \item We propose a framework for measuring a model's understanding of causal graphs by defining four specific criteria.
    \item We construct \texttt{CLEAR}, the first benchmark designed specifically to assess language models' understanding of causal graphs. \texttt{CLEAR} features three levels, encompasses 20 causal tasks, and considers six question types.
    \item Extensive experiments with six leading language models yield insightful findings and valuable observations about their capacity for understanding causal graphs.
\end{itemize}

\begin{figure*}[t!]
\centering  
\includegraphics[width=.9\textwidth]{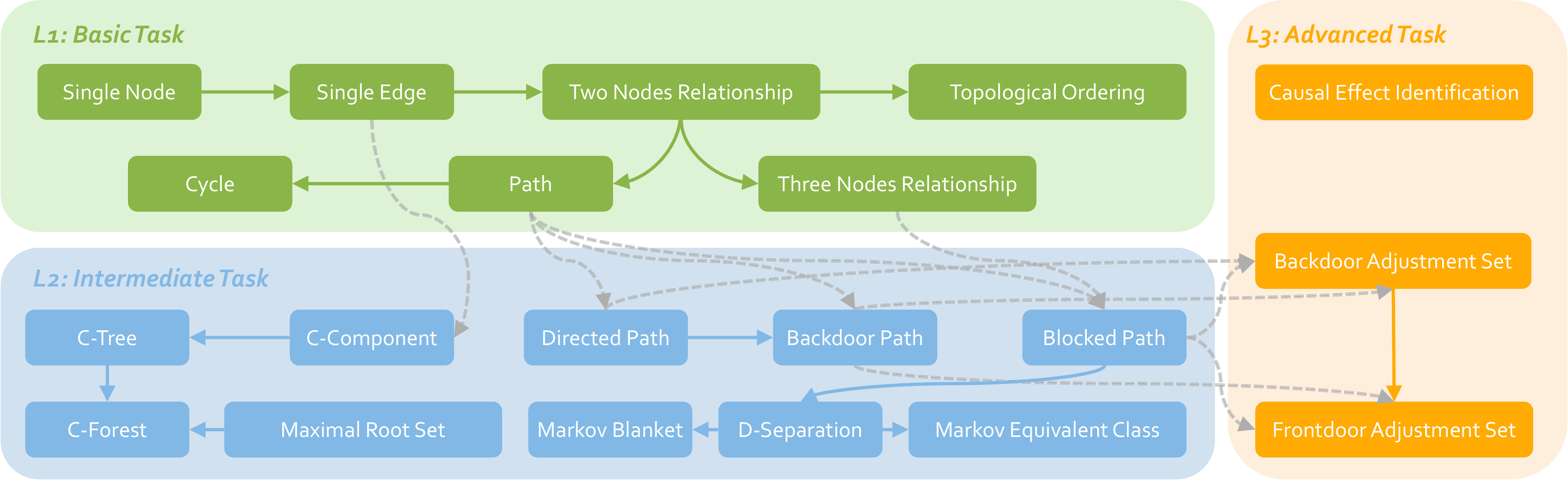}
\caption{\textbf{Hierarchy and dependent relationships of tasks in \texttt{CLEAR}.} We define three complexity levels. (1) Level 1: \textit{Basic Task}. Mastering these concepts is a prerequisite for understanding any general graph. (2) Level 2: \textit{Intermediate Task}. These tasks represent the most common characteristics in causal graphs. Causal graph-based reasoning relies heavily on understanding these fundamental problems. (3) Level 3: \textit{Advanced Task}. These tasks present complex, high-level challenges that are central to causal graph understanding. Solid arrows indicate the dependencies between tasks within the same level, while dashed arrows represent the tasks' dependencies across different levels. Task dependency design draws on established research \citep{shpitser2006identification,pearl2009causality,bareinboim2012causal,pearl2016causal,pearl2018book,jaber2019causal}.
}
\label{fig_sec1:causal_task_relation}
\end{figure*}

\section{What Do We Mean by \emph{Understanding} in Language Models?}
\label{sec:understand}

\subsection{Multiple Facets of \emph{Understanding}}
\label{subsec:understand_facet}

Unlocking the mysteries of human social behavior \citep{adler2006understanding}, decision-making \citep{frensch2014definitions}, and personality development \citep{lapsley2004moral} hinges on our ability of \emph{understanding}.  
Our investigation into \emph{understanding} begins with a brief summary of the existing definitions across various disciplines.

From the philosophical and psychological perspectives, \emph{understanding} means: 
(1) More than just knowing isolated facts. 
    It involves recognizing and grasping the relationships that weave together the various elements of a subject \citep{kvanvig2003value,carter2014objectual,sep-understanding}.
(2) Beyond the formula or definition. 
    It encompasses the ability to not only grasp concepts or formulas but also to adeptly apply them in practical contexts \citep{rumelhart1991understanding,de2004discussion}.
(3) Variation in degree. 
    Understanding is not binary, its completeness depends on the individual's conceptual context and background knowledge \citep{nickerson1985understanding}.

Considering recent machine learning endeavors, \citet{choudhury2022machine} propose three criteria to assess if a reading comprehension model reaches human-level ability. They focus on whether a model could solve problems correctly, whether it uses information that humans would deem relevant, and whether its performance is consistently robust. Although the three conditions provided in \citet{choudhury2022machine} sufficiently define a model's \emph{understanding}, there is still room for improvement. For instance, these conditions fail to offer precise quantitative criteria and lack explicit clarification on what type of information is considered relevant.

\subsection{Exploring Language Models' \emph{Understanding} of Causal Graphs}
\label{subsec:understand_explor}

Numerous studies have identified understanding as a key factor in the pursuit of human-level artificial intelligence \citep{mccarthy2007here,adams2012mapping,mcclelland2020placing}. However, arriving at a definition of \emph{understanding} within language models is an ongoing challenge. 
Evaluating models' \emph{understanding} based on accuracy is currently the dominant approach and certainly essential \citep{ashwani2024cause,he2024can,xu2024good}, but this method suffers from inherent limitations. Real-world problems are complex, and data often contains noise \citep{gupta2019dealing,moran2020noisier2noise,bansal2022data}. These make it practically impossible for any model to be perfectly accurate all the time (even humans rarely achieve this) \citep{valverde2014100}. While it is clear that understanding varies in degree \citep{nickerson1985understanding}, pinning down a specific threshold is difficult. This difficulty is compounded by the variability in task complexity and the subjective nature of interpreting ``levels of understanding''. Consequently, rather than define \emph{``what constitutes understanding of causal graphs in a language model''}, we think it might be more productive to consider \emph{``if a language model understands causal graphs, how should it behave?''}

\subsection{Seeking \emph{Understanding} of Causal Graphs in Model Behavior}
\label{subsec:understand_behavior}

To measure how well language models understand causal graphs, we develop a three-level evaluation hierarchy comprising 20 meticulously crafted causal graph-based tasks (as Figure \ref{fig_sec1:causal_task_relation} illustrates). These tasks include graphs' basic tasks (e.g., cycle), intermediate tasks (e.g., markov equivalent class), and advanced tasks (e.g., causal effect identification). Proficiency in these 20 tasks serves as a valid measure of a model's understanding of causal graphs. Therefore, combining the analyses from \cref{subsec:understand_facet} and \cref{subsec:understand_explor}, we propose that a language model that exhibits understanding would demonstrate the following four behaviors in our tasks.\footnote{More thoughts about our framework are in \cref{appendix_framework_design}.} The performance of a model is denoted by $P$, random guess by $P_r$, the original response of a model by $R$, and the ground truth by $GT$.

\paragraph{B1: Performance exceeding random guesses.} 
Existing work suggests that random guess implies a lack of extensive understanding of the given problem \citep{capraro2012investigation}. Moreover, using random guess as baseline is a common and reasonable practice in evaluating model performance \citep{chen2023theoremqa,wang2024can,chen2024causal}. This behavior can be formulated as $P>P_r$.

\paragraph{B2: Robustness against question types.}
Numerous studies highlight that altering the question type or description of a graph, while preserving the original meaning of the problem, can significantly impact model performance \citep{fatemi2023talk,hu2023beyond,luo2024graphinstruct}. Therefore, we suppose that if a model's \emph{understanding} of a causal graph and its related tasks is genuine, its performance should not be sensitive to superficial changes in the causal graph's question type. 

\paragraph{B3: Correct utilization of causal definitions.}
As \citet{de2004discussion} emphasizes, understanding implies the ability to utilize given definitions to solve problems. This behavior indicates that the model not only recognizes terms but also understands their meanings and how they relate to the given context. This behavior can be defined as: $R\leftarrow def.=GT$, where $R\leftarrow def.$ means a model's response after adding a causal definition to the prompt. The definition can be conveyed either explicitly within the prompt or implicitly through the provision of examples (e.g., IcL) \citep{li2022emergent,zheng2023step,richens2024robust}.

\paragraph{B4: Performance constrained by task dependence.}
Task dependence consistently emerges as a crucial factor in studies focused on understanding \citep{kvanvig2003value,carter2014objectual,sep-understanding}. As shown in Figure \ref{fig_sec1:causal_task_relation}, we determine that Task B is dependent on Task A if it requires knowledge acquired from Task A for resolution, whereas solving Task A does not necessitate knowledge from Task B. Mastery of the foundational task is thus deemed essential for succeeding in the dependent task. This performance constraint due to task dependence serves as a critical metric for assessing a model's depth of understanding.

\section{The \texttt{CLEAR} Benchmark}
\label{sec:benchmark}
To explore the question: \emph{\textbf{C}an \textbf{L}anguage mod\textbf{E}ls re\textbf{A}lly unde\textbf{R}stand causal graphs}? we propose \texttt{CLEAR}, the first benchmark dedicated to causal graph understanding. We ensure dataset diversity by accounting for various factors: the size, type, and density of causal graphs, as well as the richness of tasks and question types.

\subsection{Benchmark Construction}
\label{subsec:graph_generation}
\paragraph{Generating random graphs.}
We begin by randomly creating a set of graphs. A graph is denoted as $\mathcal{G}=(\mathcal{V},\mathcal{E})$, where $\mathcal{V}$ and $\mathcal{E}$ represent set of nodes and edges. To ensure diversity, we cover both general and causal graphs, differentiated by structure into four types: undirected graph, directed graph, directed acyclic graph (DAG), and acyclic directed mixed graph (ADMG) \citep{peters2017elements}. The undirected and directed graphs, typical of general graph types, are employed primarily in basic tasks. Conversely, DAGs and ADMGs, which are causal in nature, are utilized in intermediate and advanced tasks. To control the complexity, we vary the number of nodes ($n_v$) from 4 to 9 and adjust the number of edges ($n_e$) from $n_v-1$ to 10 for each $n_v$. These graph types involve three types of edges: undirected edge, directed edge, and bi-directed edge. The undirected edges symbolize reciprocal relationships, while the bi-directed edges suggest the presence of confounding between nodes. For ADMGs that contain both directed and bi-directed edges, we maintain the ratio of $\mbox{bi-directed}$ to $\mbox{directed}$ edges at or below $0.5$ to prevent excessive complexity. We denote nodes using letters, and to ensure neutrality and mitigate bias from the model’s potential prior knowledge, the alphabetical order of $\mathcal{V}$ is randomized.

\paragraph{Generating causal reasoning questions.}
\begin{figure}[t!]
\centering  
\includegraphics[width=.49\textwidth]{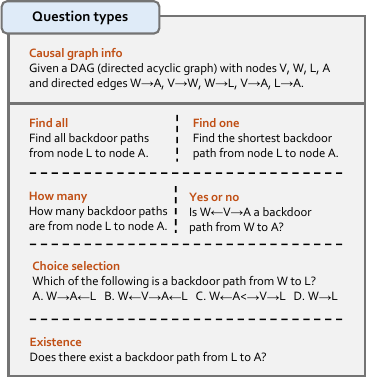}
\caption{\textbf{Six question types.} Taking the backdoor path as an example, we design six question types in \texttt{CLEAR}. A complete question is formulated by combining the causal graph info with a specific question type.}
\label{fig:question_types}
\end{figure}
Based on the causal graphs, we generate questions with corresponding ground truth for various causal tasks and question types. The questions are produced using predefined templates.
Specifically, we design 20 causal tasks and six question types.\footnote{\cref{appendix_design_detail} provides more information on the dataset.} 
And as Figure \ref{fig:question_types} demonstrates, these question types can be divided into two types of subjective questions (i.e., \emph{find all} and \emph{find one}) and four types of objective questions (i.e., \emph{how many}, \emph{yes or no}, \emph{choice selection}, and \emph{existence}), providing a in-depth evaluation of models' \emph{understanding}. 

\subsection{Data Statistics}
\label{subsec:statistics}

Our \texttt{CLEAR} benchmark includes 20 causal tasks, spanning all three complexity levels.
We generate 2808 questions in total.
For each causal task, we ensure that the number of questions exceeds 100 to support the validity of our experimental conclusions. Table \ref{table_data_statistics_concise} presents the overview of the \texttt{CLEAR}. 
\begin{center}
\begin{table}[t]
\fontsize{9}{10}\selectfont
    \caption[Statistics of the \texttt{CLEAR} benchmark.]{\textbf{Concise statistics of the \texttt{CLEAR} benchmark.} We tally the number of different causal tasks, organizing them by various levels. Type indicates question type.}
    \label{table_data_statistics_concise}
    \centering
  \begin{tabular}{l|c|c}
\toprule
\textbf{Causal task} & \textbf{\# Type}&\textbf{\# Sample}\\
\hline
\multicolumn{3}{c}{\cellcolor{green!10}\emph{Basic Task}}\\
\hline

Single node (SN)& 4&192\\
Single edge (SE)& 4&192\\
Two nodes relationship (2NR)& 5&120\\
Three nodes relationship (3NR)& 5&120\\
Path (PT)& 5&168\\
Cycle (CL)& 4&144\\
Topological ordering (TO)& 3&144\\
\hline
\multicolumn{3}{c}{\cellcolor{teal!10}\emph{Intermediate Task}}\\
\hline

Blocked path (BLP)& 3&144\\
D-separation (DS)& 3&120\\
Markov equivalent class (MEC)& 2&120\\
Markov blanket (MB)& 3&144\\
Directed path (DP)& 5&120\\
Backdoor path (BKP)& 5&144\\
C-component (CC)& 3&108\\
C-tree (CT)& 1&120\\
C-forest (CF)& 1&120\\
Maximal root set (MRS)& 4&192\\
\hline
\multicolumn{3}{c}{\cellcolor{orange!10}\emph{Advanced Task}}\\
\hline
Backdoor adjustment set (BAS)& 4&132\\
Frontdoor adjustment set (FAS)& 4&144\\
Causal effect identification (CEI)& 1&120\\
\hline
\textbf{Total} & 6&2808\\
\hline
\end{tabular}
\end{table}
\end{center}
\vspace{-9mm}

\section{Experiments}
\label{sec:experiment}

\subsection{Setups}
\label{subsec:setups}

\paragraph{Models.}
Our evaluation encompasses six models. This selection includes both open-access models (\intern~\citep{ying2024internlmmath}, \mixtral~\citep{jiang2024mixtral}, and \llama~\citep{touvron2023llama}), and limited-access models (\chatgpt~\citep{chatgpt2022}, \gptf~\citep{openai2023gpt4}, and \gemini~\citep{team2023gemini}). They originate from various creators and exhibit a 
spectrum of model scales. We use the default hyper-parameter settings for all models. 

\paragraph{Prompts.}
In \cref{subsec:compare_baseline}, \ref{subsec:robustness} and \ref{subsec:constrain}, we employ the basic prompt (i.e., \texttt{<question>}). In \cref{subsec:concept}, we adopt basic prompt, 1/3-shot IcL \citep{brown2020language}, and definition-guided prompt (i.e., \texttt{<instruction, definition, question>}).\footnote{See \cref{sec:appendix_experiments} for details on these prompts.} 

\paragraph{Metrics.}
The evaluation metric is accuracy. Objective questions are assessed via answer extraction using \gptf~and exact-match scoring.\footnote{Prior studies have shown that strong language models (e.g., \gptf) can be effective judges \citep{lu2023mathvista,zheng2024judging}, demonstrating the validity of this approach.} Subjective questions are evaluated manually.

\subsection{Comparison with Random Guess}
\label{subsec:compare_baseline}
\begin{figure*}[t]
\centering  
\includegraphics[width=1.05\textwidth]{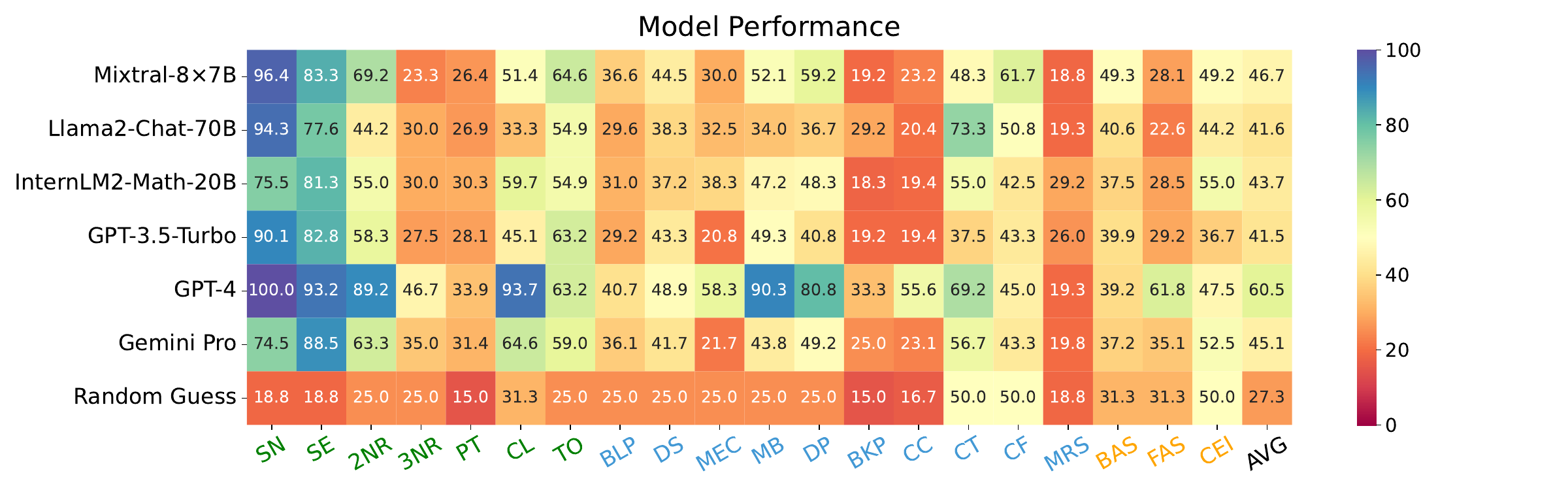}
\caption{\textbf{Overall model performance.} Each cell corresponds to the model's accuracy on that specific task.}
\label{fig:overall_performance}
\end{figure*}

Figure \ref{fig:overall_performance} illustrates the models' performances on all causal tasks. Each cell in the figure represents a model's accuracy. 
From Figure \ref{fig:overall_performance}, we can conclude that:
(1) Although limited (i.e., approximately 40\% to 60\% room for improvement), language models do exhibit preliminary \emph{understanding} (i.e., exceed random guess) of causal graphs. 
The rightmost column of the figure indicates the models' average accuracies, demonstrating that all models outperform their random guesses. This suggests that they possess basic \emph{understanding} of the causal graphs. Despite exceeding random guesses, there remains substantial room for improvement. Even the top-performing model, \gptf, only reaches an accuracy of 60.5\%, while the remaining models hover around 40.0\%. 
(2) Language models demonstrate a good grasp of the fundamental elements that constitute a causal graph. All models achieve over 70.0\% accuracy on the single node and single edge tasks, with \gptf~even reaching 100.0\% on the single node. These results provide valuable insights for designing future tasks involving causal graphs. 
(3) The model's error response is the dominant factor contributing to its subpar performance compared with random guess. We adopt the error types defined in \citet{chen2024causal} and observe the model exhibit errors such as causal hallucination, contradiction, and misunderstanding.\footnote{The qualitative analysis is provided in \cref{appendix_error_analysis}.}

Figure \ref{fig:overall_level_performance} presents the models' average accuracies across three levels. We find that:
(1) Language models excel at the \emph{basic task} level. All models achieve an accuracy exceeding 50.0\%, with the highest reaching 74.3\%. Conversely, most average accuracies attained on the remaining two levels fail to surpass 40.0\%.  
(2) The five models, excluding \gptf, demonstrate similar performance.

\begin{figure}[t!]
\centering  
\includegraphics[width=.5\textwidth]{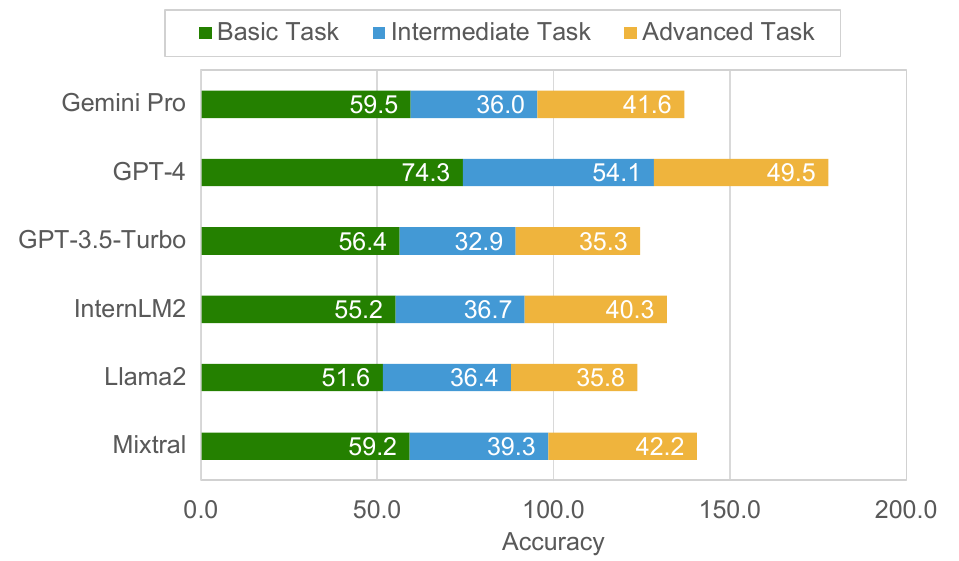}
\caption{\textbf{Model performance across the three levels of \texttt{CLEAR}.} The term Mixtral refers to \mixtral, Llama2 to \llama, and InternLM2 to \intern.}
\label{fig:overall_level_performance}
\end{figure}

\subsection{Is the Model Robust?}
\label{subsec:robustness}

To evaluate the models' robustness, we consider all six different question types. Different question types within a specific causal task, when presented with the same causal graph, aiming to probe the same core concept of causality. 
Importantly, we acknowledge the potential impact of question type on both the probabilities of random guesses and the phrasing of questions. Our objective is to conduct a preliminary investigation into how question types influence model robustness.

\begin{figure}[t!]
\centering  
\includegraphics[width=.46\textwidth]{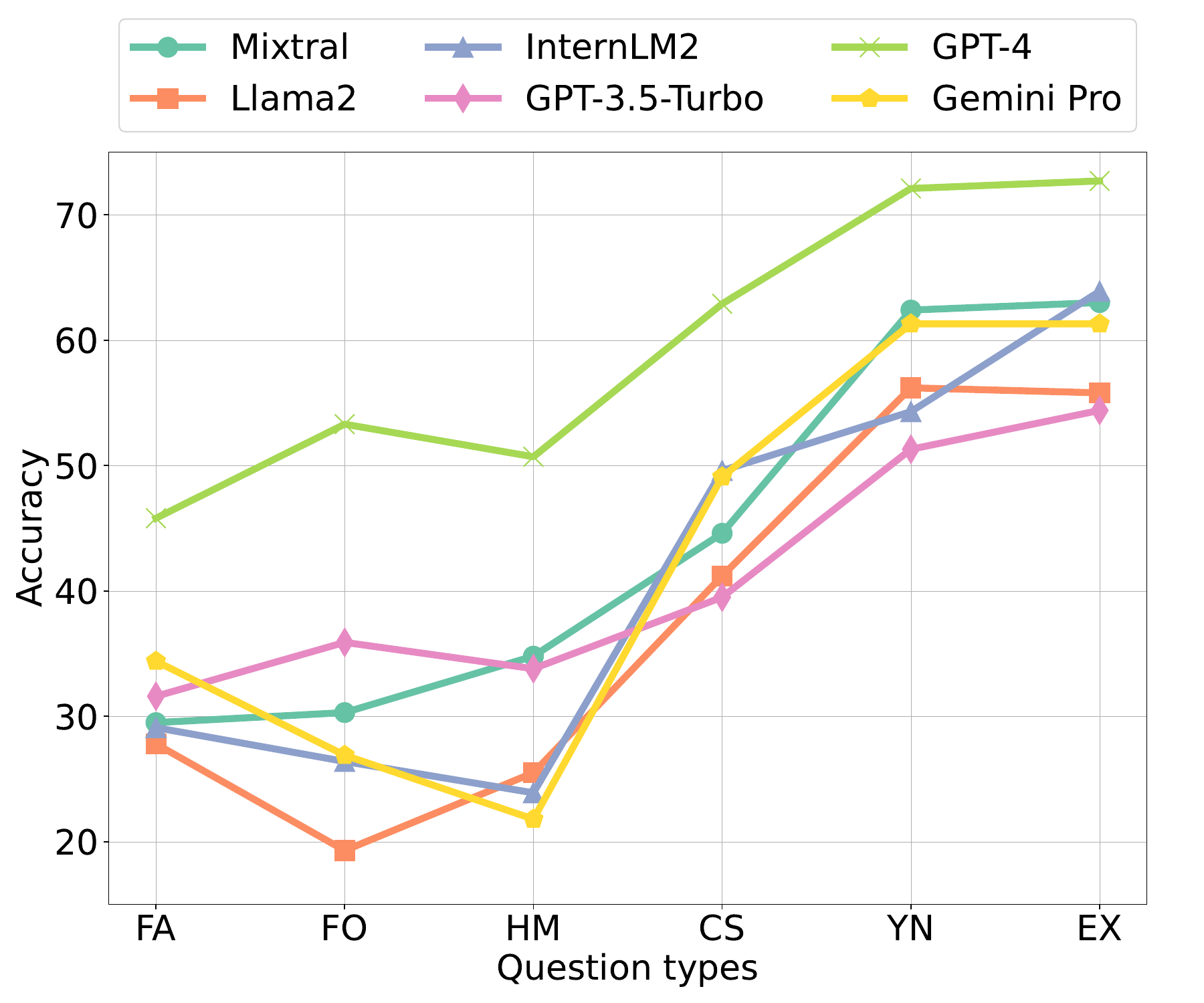}
\caption{\textbf{How the question types affect model robustness.} We compare the models' accuracies across different question types. FA stands for find all, FO for find one, HM for how many, CS for choice selection, YN for yes or no, and EX for existence.}
\label{fig:robustness_question_types}
\end{figure}

Figure \ref{fig:robustness_question_types} shows the average accuracy of the models across different question types. We draw the following conclusions:
(1) Model performance is sensitive to question type. All models excel in YN and EX question types but struggle with FA, FO, and HM. Wherein, \llama, \intern, and \gemini~exhibit performance discrepancies exceeding 35.0\% across different question types. Although \chatgpt~is not the top performer, it demonstrates the minimal performance difference, measuring at 22.8\%.
(2) A model’s understanding of causal graphs might be artificially inflated if evaluation relies on limited question types. The selection bias inherent in language models raises concerns about their robustness \citep{zheng2024large,chen2024quantifying}. If we only evaluate language models on CS, YN and EX, we risk overestimating their true capabilities. It is the diversity of question types that reveals the actual understanding capability of a model.\footnote{There is a growing trend in benchmark design towards incorporating a wider variety of question types or providing more choices for models \citep{chen2024causal,wang2024mmlu,rottger2024political}.}

\subsection{Definition Proficiency of the Model}
\label{subsec:concept}
To investigate whether the models could effectively utilize the provided definitions related to a causal graph, we further conduct experiments on seven tasks (i.e., 3NR, PT, BLP, BKP, CC, MRS, and FAS).\footnote{These abbreviations are given in Table \ref{table_data_statistics_concise}. Detailed definitions of all seven tasks are in Table \ref{table_all_definitions} (\cref{appendix_definition_proficiency}).}
For these tasks, the average accuracies across all models on the objective questions are below 40\%. Moreover, the seven tasks span all levels in Figure \ref{fig_sec1:causal_task_relation}, which can fully demonstrate the effectiveness of the experiments. For prompts, we select the basic prompt, 1/3-shot IcL, and definition-guided prompt. There is ample work validating the effectiveness of IcL \citep{wu2023self,wang2023self}. Therefore, to assess a model's ability to correctly apply or abstract a causal definition, IcL serves as an ideal reference.

\begin{figure}[t!]
\centering  
\includegraphics[width=.5\textwidth]{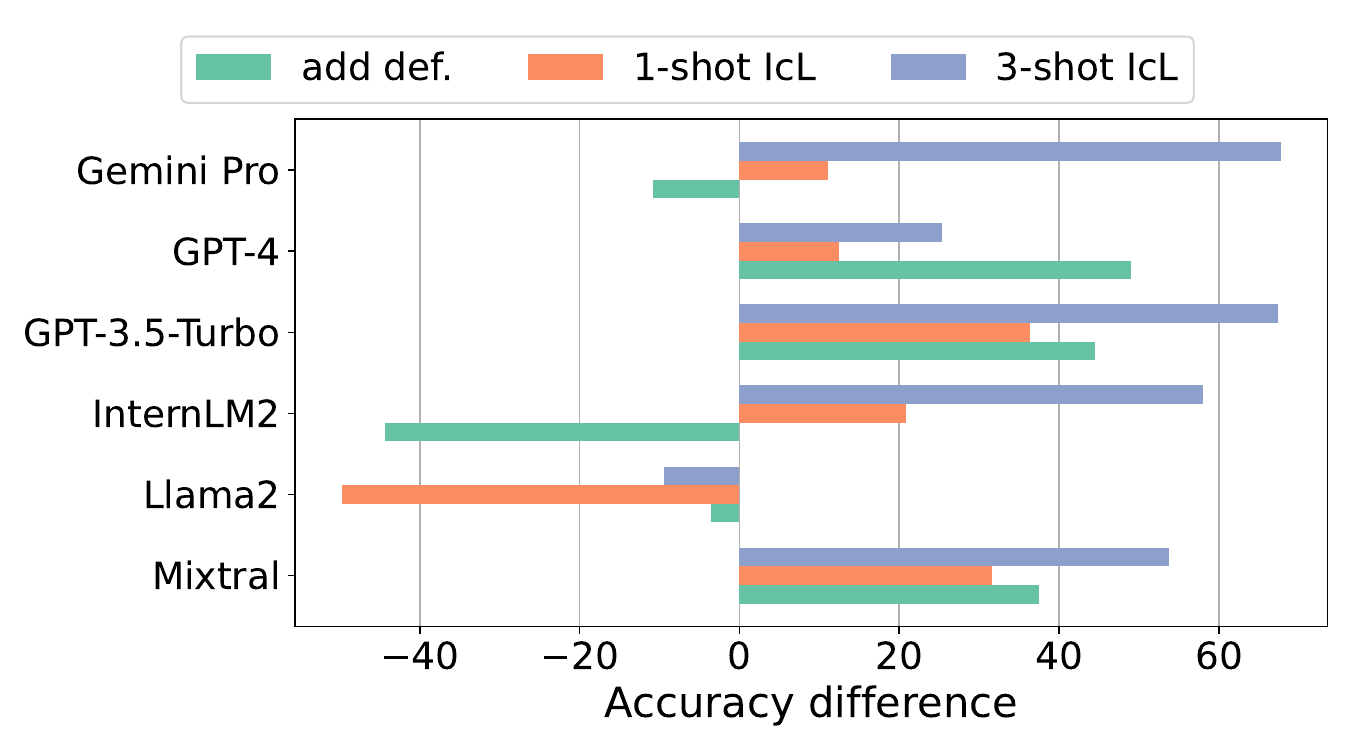}
\caption{\textbf{Explicit and implicit definition proficiency.} We compare how well the model could utilize definitions, examining both explicitly and implicitly. Add def. indicates the definition-guided prompt.}
\label{fig:definition_IcL_overall}
\end{figure}

Figure \ref{fig:definition_IcL_overall} shows the overall accuracy difference of each model across seven causal tasks using different prompts.\footnote{We provide the detailed data in Table \ref{table_definition_IcL} (\cref{appendix_definition_proficiency}).} The baseline for comparison is the average accuracy of each model under the basic prompt. By analyzing this figure, we can draw the following conclusions: 
(1) The models exhibit notable differences in their \emph{understanding} of definitions related to a causal graph. Providing the causal definition significantly enhances the performance of \gptf, \chatgpt~and \mixtral. Notably, the improvement is most pronounced for \gptf, which even surpasses both 1-shot IcL and 3-shot IcL. The improvements on \chatgpt~and \mixtral~are also remarkable, both outperforming 1-shot IcL. However, the remaining three models do not benefit from the provided definition. Specifically, \intern~exhibits the most prominent accuracy decline. 
(2) Models capable of (explicitly) utilizing definitions correctly are often observed performance improvements (implicitly) through IcL. However, even if a model's performance can be considerably promoted by IcL, it does not necessarily mean the model can successfully apply (explicit) definitions. Despite the potential for accuracy gains (over 60\% cumulatively) of \gemini~using 3-shot IcL, it struggles to correctly apply (explicitly) provided definitions, resulting in diminished performance.

\subsection{How Task Dependence Shapes Model Performance}
\label{subsec:constrain}

\begin{figure*}[t!]
\centering  
\subfigure[CC$\rightarrow$CT$\rightarrow$CF]{ 
\begin{minipage}{5cm}
\centering    
    \includegraphics[width=\linewidth]{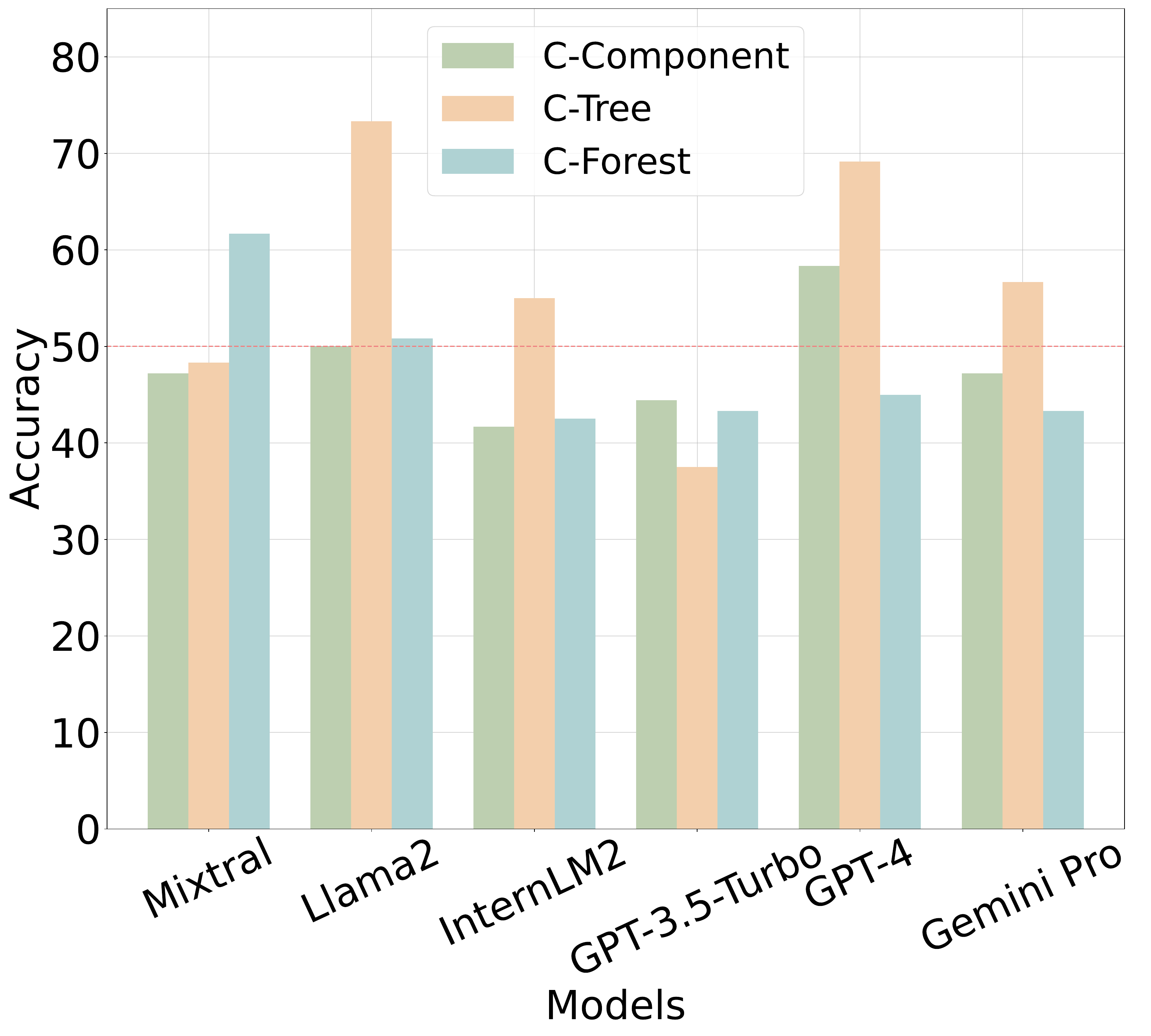}
\end{minipage}
}
\subfigure[3NR$\rightarrow$BKP$\rightarrow$BAS]{ 
\begin{minipage}{5cm}
\centering    
    \includegraphics[width=\linewidth]{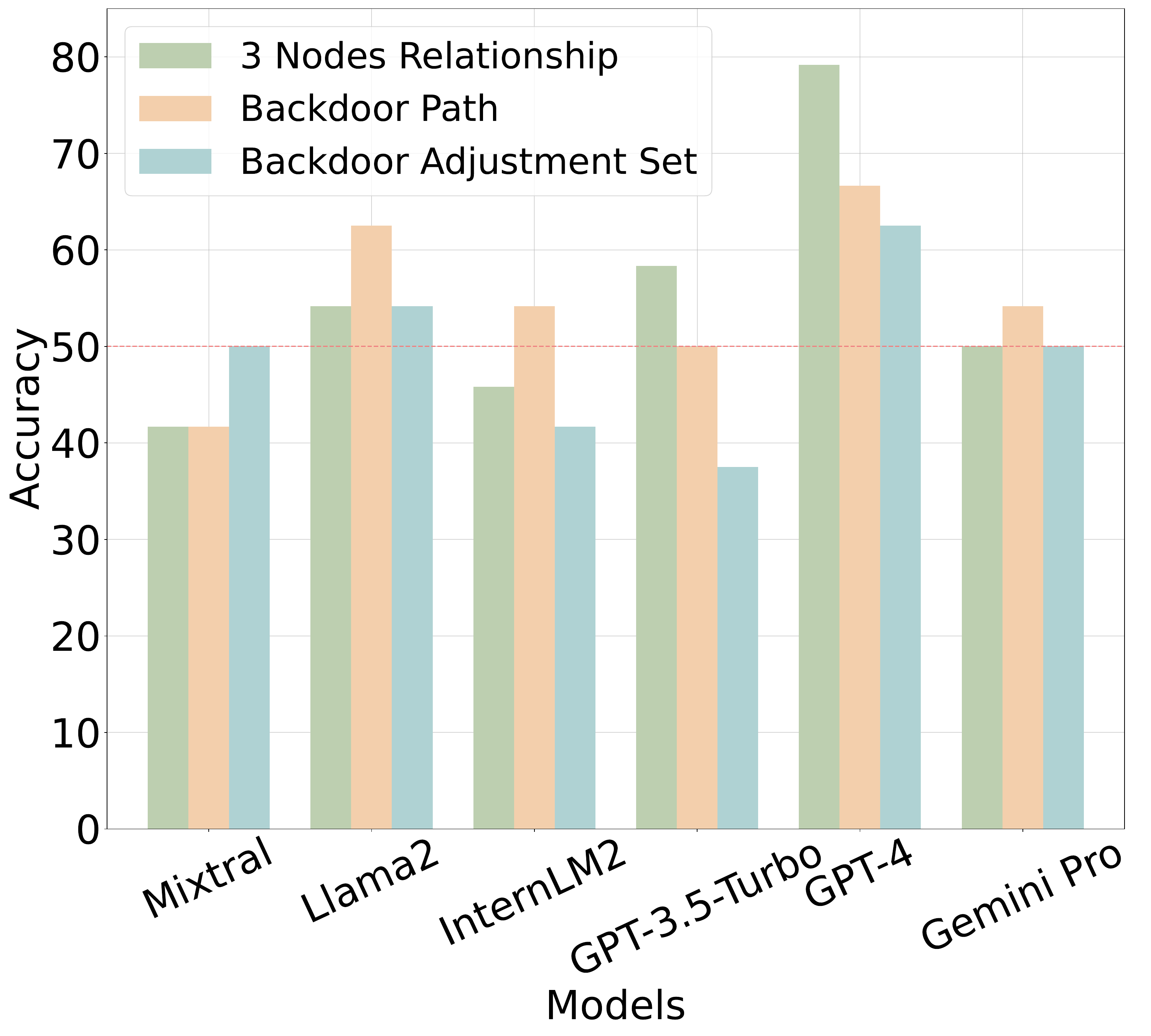}
\end{minipage}
}
\subfigure[3NR$\rightarrow$BLP$\rightarrow$DS]{ 
\begin{minipage}{5cm}
\centering    
    \includegraphics[width=\linewidth]{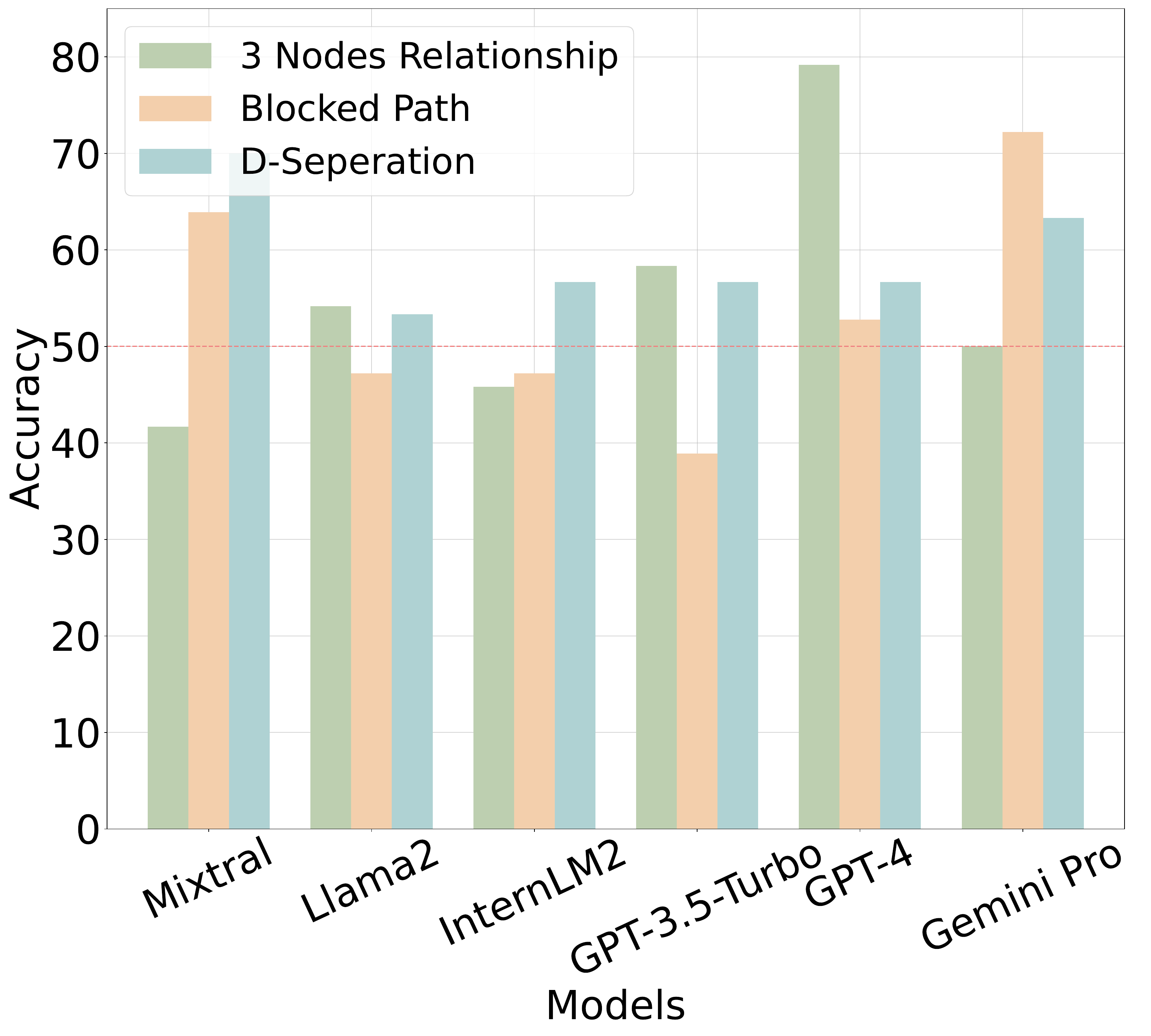}
\end{minipage}
}
\caption[How task relevance shapes model performance]{\textbf{Task dependence's impact on model performance.} We evaluate model performance across three groups of causal tasks categorized by their correlations. The orange dashed line represents the accuracy of random guess.}    
\label{fig:task_performance_constrain}    
\end{figure*}

Based on Figure \ref{fig_sec1:causal_task_relation}, we select three representative sets of dependent causal tasks and consider the YN question type. 
(1) \textit{Tasks within the same level}: we choose CC$\rightarrow$CT$\rightarrow$CF, all located at \emph{intermediate task}.
(2) \textit{Tasks across distinct levels}: we choose a sequence spanning different levels: 
3NR (\emph{basic task})$\rightarrow$BKP (\emph{intermediate task})$\rightarrow$BAS (\emph{advanced task}) are considered.
(3) \textit{Tasks with partial level overlap}: we focus on a combination where some tasks share the same level: 3NR (\emph{basic task})$\rightarrow$BKP (\emph{intermediate task})$\rightarrow$DS (\emph{intermediate task}).

Upon meticulous examination of Figure \ref{fig:task_performance_constrain}, we have the following observations:
(1) The performances of most models are not constrained by dependent causal tasks. Out of all models, only \chatgpt~and \gptf~in Figure \ref{fig:task_performance_constrain}(b) exhibit the expected accuracy trend (i.e., 3NR$\geq$BKP$\geq$BAS). These results suggest that the models might not truly \emph{understand} the causal relationships between tasks, but rather rely on other spurious correlations. It is also possible that not all models possess the capacity for human-level causal reasoning and knowledge transfer ability.
(2) Different models exhibit varying performance trends when tackling the same group of dependent causal tasks. This highlights the heterogeneity of knowledge representation and application among different models.

\subsection{Counterfactual Explainability}
\label{subsec:counterfactual}

\begin{figure}[t!]
\centering  
\includegraphics[width=.46\textwidth]{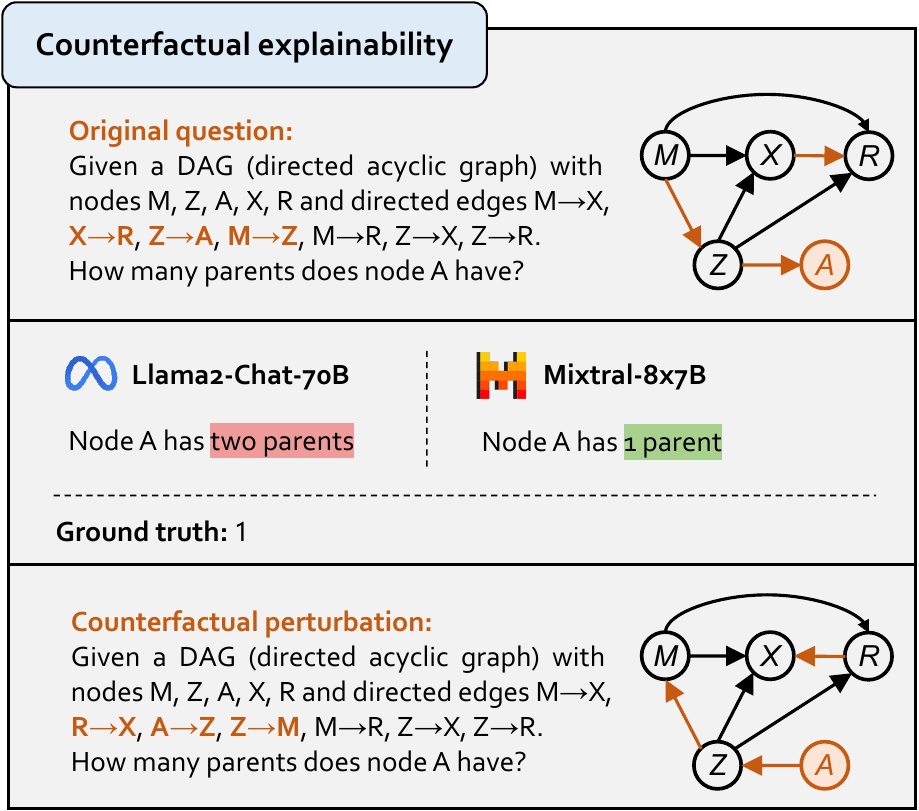}
\caption{\textbf{Counterfactual perturbation used in this case.} Starting with the original question, we obtain answers for both models. Next, we establish the baseline using counterfactual perturbation. Finally, we calculate the token attribution of \textcolor{orange}{key information} to understand its influence on the model's output.}
\label{fig:example_counterfactual}
\end{figure}

The analyses from \cref{subsec:compare_baseline} to \cref{subsec:constrain} are based on directly calculating the accuracy of the models' outputs. To extend beyond mere accuracy, we leverage \texttt{Captum} \citep{kokhlikyan2020captum,miglani2023using}, a \texttt{Python} library for model interpretability, to explore language models' \emph{understanding} of causal graphs from a counterfactual perspective. We primarily use the \emph{perturbation based methods} provided by \texttt{Captum}.\footnote{For further guidance, refer to the tutorial at: \url{https://captum.ai/tutorials/Llama2_LLM_Attribution}.} As depicted in Figure \ref{fig:example_counterfactual}, we first query both \llama~and \mixtral, which are of comparable scale and have been widely adopted, using the original question to obtain their respective answers. Our main focus is the impact of ``Z$\rightarrow$A'' on the model's response. We suspect ``X$\rightarrow$R'' and ``M$\rightarrow$Z'', which are located near the ``Z$\rightarrow$A'', could also potentially impact the model's response. Consequently, we use counterfactual perturbations to analyze the influences of these three statements on the model. We set the counterfactual perturbations as baseline (i.e., perturbation-based algorithm uses it as reference value), and the model's response as target string. Finally, using the target function in \texttt{Captum}, we calculate the log probability of the model generating its answer given the question. 

\begin{figure}[t!]
\centering  
\includegraphics[width=.5\textwidth]{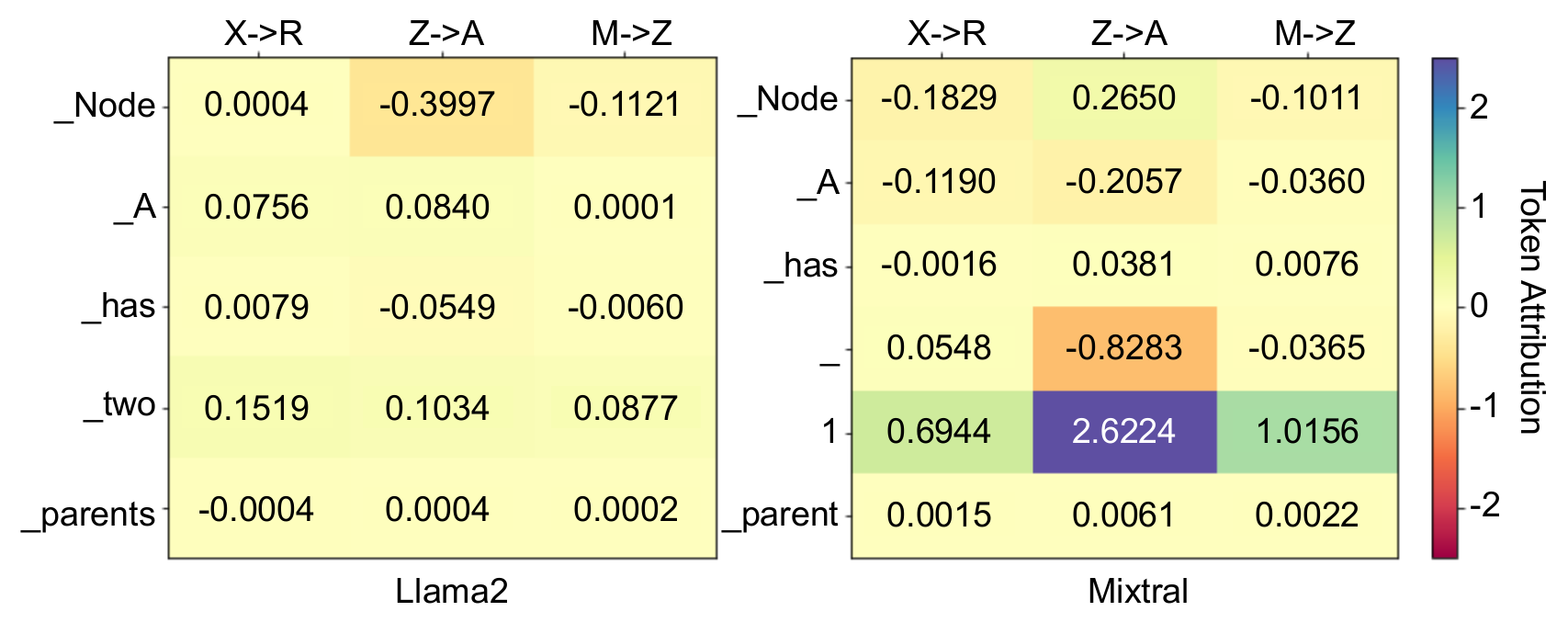}
\caption{\textbf{Token attribution.} On the y-axis, underscores mark the tokenizer's divisions of each target string. The x-axis displays key information of the question.}
\label{fig:counterfactual}
\end{figure}

Figure \ref{fig:counterfactual} displays the token attributions of the models' responses. We find that ``Z$\rightarrow$A'' is the most positive factor in getting the right answer ``1'' for \mixtral, with ``X$\rightarrow$R'' and ``M$\rightarrow$Z'' also contributing positively. This confirms that \mixtral~correctly identifies and utilizes the relevant information in its reasoning process. In contrast, \llama~produces a wrong answer. Neither ``Z$\rightarrow$A'', ``X$\rightarrow$R'' nor ``M$\rightarrow$Z'' exhibit a significant positive impact on its answer, suggesting that \llama~fails to identify key information. The results support the claim in \cref{subsec:compare_baseline} that models have a preliminary understanding of causal graphs. More importantly, the results demonstrate a strong link between a model's \emph{understanding} of the causal graph and its ability to focus on the essential information within the graph. 

\section{Related Work}

\paragraph{Language models' \emph{understanding} ability.}
Language models' understanding is being probed through various perspectives, such as causality \citep{hobbhahn2022investigating,kim2023can,ashwani2024cause}, real-world problems \citep{choi2023llms,he2024can,xu2024good}, disciplines \citep{castro2023large,guo2024learning}. A common approach in these studies is to establish a benchmark, and then evaluate a model's performance. 
A more rigorous exploration of what means \emph{understanding} in models is still needed.

\paragraph{Graph-based benchmarks.}
The capacity of language models to solve graph-based problems is attracting growing attention. \citet{wang2024can} propose the NLGraph, concentrating primarily on essential graph tasks. \citet{luo2024graphinstruct} introduce the GraphInstruct benchmark. \citet{fatemi2023talk} propose the GraphQA to explore the impact of different graph encoding methods. LLM4DyG \citep{zhang2023llm4dyg} addresses the dynamic graphs. Despite progress in applying models to graph tasks, their ability to reason about causality within graphs still requires further investigation.

\section{Conclusion}
This paper provides a comprehensive and in-depth exploration on the question: \emph{Can language models really understand causal graphs?} We define a practical framework for accessing a model's understanding. We introduce \texttt{CLEAR}, a novel benchmark designed to evaluate a model's understanding of causal graphs, filling a significant gap in existing research. We validate our framework through extensive experiments and conclude five insightful findings. 

\section{Limitations}
Despite our best efforts to design a framework for causal graph understanding, construct a benchmark, and conduct thorough experiments on six models, we acknowledge that our work still has limitations. 
The language of \texttt{CLEAR} is relatively limited. 
Due to time and budget constraints, our benchmark only considers English. As language models are increasingly used worldwide, we acknowledge that a multilingual dataset could provide more meaningful findings.
Moreover, the definition of understanding still requires further exploration. For instance, how to extend the concept of robustness to broader scenarios. Additionally, evaluating the understanding of large vision language models (LVLMs) will likely require considering a wider set of factors.

\bibliography{custom}

\clearpage
\appendix

\section{Considerations in Understanding Framework Design}
\subsection{Two-pronged Approach}
\label{appendix_framework_design}

In \cref{subsec:understand_explor}, we propose that we should consider \emph{``if a language model understands causal graphs, how should it behave?''} We believe the question needs to be considered from two aspects: (1) Human-centric perspective. To properly assess language models, we must first define understanding in a way that aligns with human cognition. This is crucial for ensuring language models truly achieve human capabilities. (2) Model-centric perspective. While human-centric definitions provide a starting point, there exist foundational differences in information processing between human brains and language models \citep{caucheteux2023evidence}. Therefore, we need to explore practical definitions that are suitable to the characteristics of models. 

To this end,  we carefully design four criteria in \cref{subsec:understand_behavior}. And from a model-centric perspective, we define understanding by examining behaviors related to B1: performance exceeding random guesses and B2: robustness against question types. From a human-centric perspective, we consider B3: correct utilization of causal definitions and B4: performance is constrained by task dependence.

\section{Design Details of \texttt{CLEAR}}
\label{appendix_design_detail}

\subsection{Overall Statistics}
\label{appendix_statistics}
The detailed statistics of \texttt{CLEAR} are in Table \ref{table_data_statistics}.

\subsection{Question Templates}
\label{appendix_templates}
Templates for questions of \texttt{CLEAR} are listed in Table \ref{table_template}. While potentially impacting diversity \citep{cobbe2021training}, this method enables efficient data scaling and accesses whether a model can recognize subtle distinctions within the templates \citep{chen2024causal}.

\begin{center}
\begin{table*}[t]
\fontsize{9}{10}\selectfont
    \caption[Statistics of the \texttt{CLEAR} benchmark.]{\textbf{Detailed statistics of the \texttt{CLEAR} benchmark.} We tally the number of different question types within each causal task, organizing them by various levels. YN indicates yes or no.}
    \label{table_data_statistics}
    \centering
  \begin{tabular}{l|c|c|c|c|c|c|c}
\toprule
\textbf{Causal task} &  \textbf{Find all}& \textbf{Find one}& \textbf{How many}&  \textbf{Choice selection}&\textbf{YN}&\textbf{Existence}&\textbf{Total}\\
\hline
\multicolumn{8}{c}{\cellcolor{green!10}\emph{Basic Task}}\\
\hline

Single node (SN)& 48& -& 48&  48&48&-&192\\
Single edge (SE)& 48&  -& 48&  48&48&-&192\\
Two nodes relationship (2NR)& 24&  -& 24&  24&24& 24&120\\
Three nodes relationship (3NR)& 24&  -& 24&  24&24& 24&120\\
Path (PT)& 24& 72& 24&  24&24&-&168\\
Cycle (CL)&-& 36&-&  36&36&36&144\\
Topological ordering (TO)&-& 48&-&  48&48&-&144\\
\hline
\multicolumn{8}{c}{\cellcolor{teal!10}\emph{Intermediate Task}}\\
\hline

Blocked path (BLP)&-&   72&-&  36&36&-&144\\
D-separation (DS)&-&   60&-&  30&30&-&120\\
Markov equivalent class (MEC)&-&   60&-&  -&60&-&120\\
Markov blanket (MB)&-& 48&-&  48&48&-&144\\
Directed path (DP)& 24&-& 24&  24&24& 24&120\\
Backdoor path (BKP)& 24& 48& 24&  24&24&-&144\\
C-component (CC)& 36&-& 36&  -&36&-&108\\
C-tree (CT)&-&-&-&  -&120&-&120\\
C-forest (CF)&-&-&-&  -&120&-&120\\
Maximal root set (MRS)& 48&-& 48&  48&48&-&192\\
\hline
\multicolumn{8}{c}{\cellcolor{orange!10}\emph{Advanced Task}}\\
\hline
Backdoor adjustment set (BAS)&-&   72&-&  24&24& 12&132\\
Frontdoor adjustment set (FAS)&-&   72&-&  24&24& 24&144\\
Causal effect identification (CEI)&-&  -&-&  -&120&-&120\\
\hline
\textbf{Total} & 300& 588& 300&  510&966&144&2808\\
\hline
\end{tabular}
\end{table*}
\end{center}

\begin{center}
\begin{table*}[t]
\fontsize{8}{9}\selectfont
    \caption{\textbf{Question template for \texttt{CLEAR}.} }
    \label{table_template}
    \centering
\begin{tabularx}{\textwidth}{l|c|X} 
\toprule
\textbf{Causal task} &  \textbf{Type}& \makecell[c]{\textbf{Template}}\\
\hline
\multicolumn{3}{c}{\cellcolor{green!10}\emph{Basic Task}}\\
\hline
\multirow{4}{*}{SN}&  FA&\texttt{List all nodes of this graph.}\\
&  HM&\texttt{How many nodes does this graph have?}\\
&  CS&\texttt{Which of the following \textcolor{orange}{is/is NOT} a node of this graph?}\\
&  YN&\texttt{Is \textcolor{orange}{\{variable\}} a node of this graph?}\\

\hline
\multirow{4}{*}{SE}&  FA&\texttt{List all edges of this graph.}\\
&  HM&\texttt{How many edges does this graph have?}\\
&  CS&\texttt{Which of the following \textcolor{orange}{is/is NOT} an edge of this graph?}\\
&  YN&\texttt{Is \textcolor{orange}{\{variable\}} a edge of this graph?}\\
\hline

\multirow{5}{*}{2NR}&  FA&\texttt{List all \textcolor{orange}{parents/descendants/children/ancestors} of \textcolor{orange}{\{variable\}}.}\\
&  HM&\texttt{How many \textcolor{orange}{parents/descendants/children/ancestors} does \textcolor{orange}{\{variable\}} have?}\\
&  CS&\texttt{Which of the following is one of \textcolor{orange}{parents/descendants/children/ancestors} of \textcolor{orange}{\{variable\}}?}\\
&  YN&\texttt{Is \textcolor{orange}{\{variable\}} one of \textcolor{orange}{parents/descendants/children/ancestors} of \textcolor{orange}{\{variable\}}?}\\
&  EX&\texttt{Does \textcolor{orange}{\{variable\}} have any \textcolor{orange}{parents/descendants/children/ancestors}?}\\
\hline
\multirow{5}{*}{3NR}&  FA&\texttt{List all \textcolor{orange}{chains/forks/v-structures} of this graph.}\\
&  HM&\texttt{How many \textcolor{orange}{chains/forks/v-structures} does this graph have?}\\
&  CS&\texttt{Which of the following is a \textcolor{orange}{chain/fork/v-structure} of this graph?}\\
&  YN&\texttt{Does \textcolor{orange}{\{variables\}} form a \textcolor{orange}{chain/fork/v-structure} in this graph?}\\
&  EX&\texttt{Are there any \textcolor{orange}{chain/fork/v-structure} of this graph?}\\
\hline

\multirow{5}{*}{PT}&  FA&\texttt{Find all path from \textcolor{orange}{\{variable\}} to \textcolor{orange}{\{variable\}}.}\\
&  FO&\texttt{Find \textcolor{orange}{one/the shortest/the longest} path from \textcolor{orange}{\{variable\}} to \textcolor{orange}{\{variable\}}.}\\
&  HM&\texttt{How many paths are there from \textcolor{orange}{\{variable\}} to \textcolor{orange}{\{variable\}}.}\\
&  CS&\texttt{Which of the following is a path from \textcolor{orange}{\{variable\}} to \textcolor{orange}{\{variable\}}?}\\
&  YN&\texttt{Is \textcolor{orange}{\{variable\}} a path from \textcolor{orange}{\{variable\}} to \textcolor{orange}{\{variable\}}?}\\
\hline

\multirow{4}{*}{CL}& FO&\texttt{Find one cycle in this graph.}\\
&  CS&\texttt{Which of the following is a cycle in this graph?}\\
&  YN&\texttt{Is \textcolor{orange}{\{variable\}} a cycle in this graph?}\\
&  EX&\texttt{Are there any cycle in this graph?}\\
\hline

\multirow{3}{*}{TO}& FO&\texttt{Find one valid topological ordering in this graph.}\\
&  CS&\texttt{Which of the following is a valid topological ordering of this graph?}\\
&  YN&\texttt{Is \textcolor{orange}{\{variable\}} a valid topological ordering of this graph?}\\
\hline

\multicolumn{3}{c}{\cellcolor{teal!10}\emph{Intermediate Task}}\\
\hline

\multirow{3}{*}{BLP}&  FO&\texttt{Find \textcolor{orange}{one valid/the minimal} nodeset that can block \textcolor{orange}{\{variable\}}.}\\
&  CS&\texttt{Which of the following nodesets can block \textcolor{orange}{\{variable\}}?}\\
&  YN&\texttt{Can \textcolor{orange}{\{variable\}} be blocked by \textcolor{orange}{\{variable\}}?}\\
\hline

\multirow{3}{*}{DS}&  FO&\texttt{Find \textcolor{orange}{one valid/the minimal} nodeset that can d-separate \textcolor{orange}{\{variable\}} and \textcolor{orange}{\{variable\}}.}\\
&  CS&\texttt{Which of the following nodesets can d-separate \textcolor{orange}{\{variable\}} and \textcolor{orange}{\{variable\}}?}\\
&  YN&\texttt{Are \textcolor{orange}{\{variable\}} and \textcolor{orange}{\{variable\}} d-separated by \textcolor{orange}{\{variable\}}?}\\
\hline

\multirow{3}{*}{MEC}&  FO&\texttt{Find another graph that belongs to the same markov equivalent class of the given graph.}\\
&  \multirow{2}{*}{YN}&\texttt{Given another DAG with nodes \textcolor{orange}{\{variable\}} and directed edges \textcolor{orange}{\{variable\}}, do these two graphs belong to the same markov equivalent class?}\\
\hline

\multirow{3}{*}{MB}&  FO&\texttt{What is the markov blanket of \textcolor{orange}{\{variable\}}.}\\
&  CS&\texttt{Which of the following is the markov blanket of \textcolor{orange}{\{variable\}}?}\\
&  YN&\texttt{Is \textcolor{orange}{\{variable\}} the markov blanket of \textcolor{orange}{\{variable\}}?}\\
\hline

\multirow{5}{*}{DP}&  FA&\texttt{Find all directed paths from \textcolor{orange}{\{variable\}} to \textcolor{orange}{\{variable\}}.}\\
&  HM&\texttt{How many directed paths are there from \textcolor{orange}{\{variable\}} to \textcolor{orange}{\{variable\}}?}\\
&  CS&\texttt{Which of the following is a directed path from \textcolor{orange}{\{variable\}} to \textcolor{orange}{\{variable\}}?}\\
&  YN&\texttt{Is \textcolor{orange}{\{variable\}} a directed path from \textcolor{orange}{\{variable\}} to \textcolor{orange}{\{variable\}}?}\\
&  EX&\texttt{Is there a directed path from \textcolor{orange}{\{variable\}} to \textcolor{orange}{\{variable\}}?}\\
\hline

\multirow{5}{*}{BKP}&  FA&\texttt{Find all backdoor paths from \textcolor{orange}{\{variable\}} to \textcolor{orange}{\{variable\}}.}\\
&  FO&\texttt{Find \textcolor{orange}{the shortest/the longest} backdoor path from \textcolor{orange}{\{variable\}} to \textcolor{orange}{\{variable\}}.}\\
&  HM&\texttt{How many backdoor paths are there from \textcolor{orange}{\{variable\}} to \textcolor{orange}{\{variable\}}.}\\
&  CS&\texttt{Which of the following is a backdoor path from \textcolor{orange}{\{variable\}} to \textcolor{orange}{\{variable\}}?}\\
&  YN&\texttt{Is \textcolor{orange}{\{variable\}} a backdoor path \textcolor{orange}{\{variable\}} to \textcolor{orange}{\{variable\}}?}\\
\hline

\multirow{5}{*}{CC}& \multirow{2}{*}{FA}&\texttt{It can be uniquely partitioned into a set C(G) of subgraphs, each a maximal C-component. Write down such partition of the given graph.}\\
&  \multirow{2}{*}{HM}&\texttt{It can be uniquely partitioned into a set C(G) of subgraphs, each a maximal C-component. How many subgraphs are there in C(G)?}\\
&  YN&\texttt{Is it a C-component??}\\
\hline

\multirow{1}{*}{CT}& YN&\texttt{Is it a C-tree?}\\
\hline

\multirow{1}{*}{CF}& YN&\texttt{Is it a C-forest?}\\
\hline

\multirow{4}{*}{MRS}&  FA&\texttt{Find the maximal root set of this graph.}\\
&  HM&\texttt{How many nodes are there in the maximal root set of this graph?}\\
&  CS&\texttt{Which of the following options is the maximal root set of this graph?}\\
&  YN&\texttt{Is \textcolor{orange}{\{variable\}} the maximal root set of this graph?}\\
\hline
\multicolumn{3}{c}{\cellcolor{orange!10}\emph{Advanced Task}}\\
\hline

\multirow{4}{*}{BAS}&  FO&\texttt{Find \textcolor{orange}{one valid/one minimal/one maximal} backdoor adjustment set for \textcolor{orange}{\{variable\}} and \textcolor{orange}{\{variable\}}.}\\
&  CS&\texttt{Which of the following sets is a valid backdoor adjustment set for \textcolor{orange}{\{variable\}} and \textcolor{orange}{\{variable\}}?}\\
&  YN&\texttt{Is \textcolor{orange}{\{variable\}} a valid backdoor adjustment set for \textcolor{orange}{\{variable\}} and \textcolor{orange}{\{variable\}}?}\\
&  EX&\texttt{Does there exist a valid backdoor adjustment set for \textcolor{orange}{\{variable\}} and \textcolor{orange}{\{variable\}}?}\\
\hline

\multirow{4}{*}{FAS}&  FO&\texttt{Find \textcolor{orange}{one valid/one minimal/one maximal} frontdoor adjustment set for \textcolor{orange}{\{variable\}} and \textcolor{orange}{\{variable\}}.}\\
&  CS&\texttt{Which of the following sets is a valid frontdoor adjustment set for \textcolor{orange}{\{variable\}} and \textcolor{orange}{\{variable\}}?}\\
&  YN&\texttt{Is \textcolor{orange}{\{variable\}} a valid frontdoor adjustment set for \textcolor{orange}{\{variable\}} and \textcolor{orange}{\{variable\}}?}\\
&  EX&\texttt{Does there exist a valid frontdoor adjustment set for \textcolor{orange}{\{variable\}} and \textcolor{orange}{\{variable\}}?}\\
\hline

\multirow{1}{*}{CEI}& YN&\texttt{Can the causal effect of \textcolor{orange}{\{variable\}} on \textcolor{orange}{\{variable\}} be identified or not?}\\
\hline
\end{tabularx}
\end{table*}
\end{center}
\vspace{-1.6cm}

\section{Details for Experiments}
\label{sec:appendix_experiments}

\subsection{Prompt Settings}
\paragraph{Basic prompt.}
Our basic prompt aligns with the definition in \citet{chen2024causal}, referring to providing only the question requiring an answer.

\begin{figure}[t!]
\centering  
\includegraphics[width=.49\textwidth]{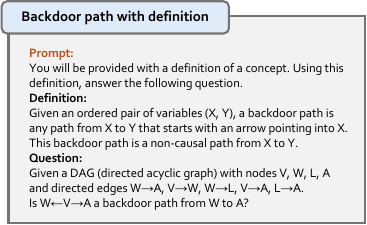}
\caption{\textbf{Definition-guided prompt.} We explicitly provide the model with definitions relevant to the questions.}
\label{fig:definition_BKP}
\end{figure}

\paragraph{Definition-guided prompt.}
Taking BKP as an example, Figure \ref{fig:definition_BKP} illustrates how to incorporate the definition of this causal task into the prompt (i.e., definition-guided prompt).

\subsection{Qualitative Analysis of Error Response}
\label{appendix_error_analysis}
\begin{figure*}[t!]
\centering  
\subfigure[Example of contradiction and misunderstanding.]{ 
\begin{minipage}{7.5cm}
\centering    
    \includegraphics[width=\linewidth]{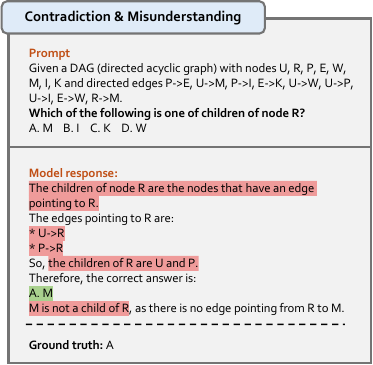}
\end{minipage}
}
\subfigure[Example of causal hallucination.]{ 
\begin{minipage}{7.5cm}
\centering    
    \includegraphics[width=\linewidth]{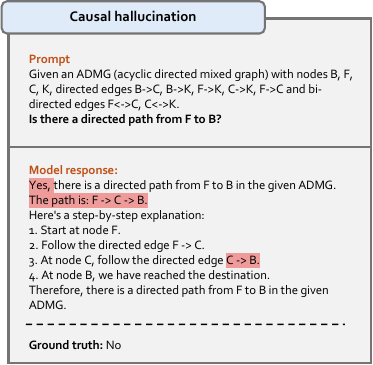}
\end{minipage}
}
\caption{\textbf{Error response.} We adopt the error types defined in \citet{chen2024causal}. The red text means the wrong response. The green text means the right response.}    
\label{fig:error}    
\end{figure*}

Figure \ref{fig:error} provides cases for models' error responses.

\subsection{Results of Definition Proficiency}
\label{appendix_definition_proficiency}
Definitions for the seven selected tasks are provided in Table \ref{table_all_definitions}.
The complete results of the four prompts are shown in Table \ref{table_definition_IcL}.
\begin{center}
\begin{table*}[t]
\fontsize{9}{10}\selectfont
\caption[Question templates]{\textbf{Definitions of the selected seven causal tasks.}}
\label{table_all_definitions}
\begin{tabularx}{\textwidth}{c|X} 
\toprule
{\textbf{Causal task}} & \makecell[c]{\textbf{Definition}} \\
\hline
\multicolumn{2}{c}{\cellcolor{green!10}\emph{Basic Task}}\\
\hline
  \multirow{8}{*}{Three nodes relationship}     &  Given a DAG with three nodes X, Y, Z. 
\newline
(1) A ``chain'' is a sequence of nodes connected by edges where each node has only one predecessor and one successor (except for the first and last nodes in the chain). The simplest chain in a causal graph can be illustrated as ``X->Y->Z''.
\newline
(2) A ``fork'' refers to a situation where one node has multiple outgoing edges leading to different successor nodes. The simplest fork in a causal graph can be illustrated as ``X<-Y->Z''.
\newline
(3) A ``v-structure'' means one node is a child of the two others that themselves are not adjacent. The simplest v-structure in a causal graph can be illustrated as ``X->Y<-Z''. \\
\hline
 \multirow{2}{*}{Path}  &A path in a DAG is a sequence of (at least two) distinct nodes $i_1,\dots,i_m$ such that there is an edge between $i_k$ and $i_{k+1}$ for all $k=1,\dots,m$. \\
\hline
\multicolumn{2}{c}{\cellcolor{teal!10}\emph{Intermediate Task}}\\
\hline
 \multirow{4}{*}{Blocked path} &In a DAG, a path p is said to be blocked by a set of nodes Z if and only if:
\newline
(1) p contains a chain i->m->j or a fork i<-m->j such that the middle node m is in Z, or 
\newline
(2) p contains an inverted fork (or collider) i->m<-j such that the middle node m is not in Z and such that no descendant of m is in Z.\\
\hline
 \multirow{2}{*}{Backdoor path} &Given an ordered pair of variables (X, Y), a backdoor path is any path from X to Y that starts with an arrow pointing into X. This backdoor path is a non-causal path from X to Y.\\
\hline
 \multirow{2}{*}{C-component} & Let G be a causal graph such that a subset of its bidirected arcs forms a spanning tree over all nodes in G. Then G is a C-component.\\
\hline 
 \multirow{2}{*}{Maximal root set} & Let G be a causal graph and X is one node that belongs to G. If X does not have any descendant, then we call X a root set of G. Maximal root set contains all the root sets of G.\\
\hline 
\multicolumn{2}{c}{\cellcolor{orange!10}\emph{Advanced Task}}\\
\hline
 \multirow{7}{*}{Frontdoor adjustment set} &If a set of variables Z satisfies the front-door criterion relative to an ordered pair of variables (X, Y):
 \newline
(1) Z intercepts all directed paths from X to Y;
\newline
(2) there is no unblocked back-door path from X to Z; and
\newline
(3) all back-door paths from Z to Y are blocked by X.
\newline
Then we call Z a frontdoor adjustment set, this set allows us to accurately estimate the causal effect of X on Y. \\
\hline
\end{tabularx}
\end{table*}
\end{center}

\begin{center}
\begin{table*}[t]
\fontsize{9}{10}\selectfont
    \caption[Statistics of the \texttt{CLEAR} benchmark.]{\textbf{Model performance on seven selected causal tasks.} }
    \label{table_definition_IcL}
    \centering
  \begin{tabular}{l|c|c|c|c|c|c|c}
\toprule
\textbf{Causal task} &  \textbf{Prompt} &\textbf{Mixtral}& \textbf{Llama2}& \textbf{InternLM2}&  \textbf{\chatgpt}&\textbf{\gptf}&\textbf{\gemini}\\
\hline
\multicolumn{8}{c}{\cellcolor{green!10}\emph{Basic Task}}\\
\hline
\multirow{4}{*}{Three nodes relationship}&  Basic&29.2& 37.5& 37.5&  34.4
&55.2
&41.7\\
&  add def.&35.4&  33.3& 29.2&  40.6&60.4&42.7\\
&  1-shot IcL&34.4&  32.3& 35.4&  40.6&57.3& 41.7\\
&  3-shot IcL&42.7&  38.5& 51.0&  44.8&54.2& 42.7\\
\hline
\multirow{4}{*}{Path}&  Basic&34.7& 36.1& 38.9&  34.7&33.3&38.9\\
&  add def.&34.7&  30.6& 33.3&  30.6&23.6&34.7\\
&  1-shot IcL&36.1&  29.2& 37.5&  26.7&31.9& 38.9\\
&  3-shot IcL&43.1&  31.9& 48.6&  44.4&26.4& 50.0\\
\hline
\multicolumn{8}{c}{\cellcolor{teal!10}\emph{Intermediate Task}}\\
\hline
\multirow{4}{*}{Blocked path}&  Basic&44.4& 40.3& 36.1&  26.4&43.1&45.8\\
&  add def.&40.3&  37.5& 18.1&  44.4&56.9&36.1\\
&  1-shot IcL&50.0&  31.9& 37.5&  44.4&44.4& 36.1\\
&  3-shot IcL&47.2&  43.1& 40.3&  43.1&48.6& 40.3\\

\hline
\multirow{4}{*}{Backdoor path}&  Basic&27.8& 43.1& 29.2&  30.6&44.4&31.9\\
&  add def.&48.6&  37.5& 18.1&  31.9&62.5&31.9\\
&  1-shot IcL&40.3&  19.4& 38.9&  36.1&52.8& 40.3\\
&  3-shot IcL&44.4&  30.6& 37.5&  40.3&55.6& 52.8\\

\hline
\multirow{4}{*}{C-component}&  Basic&29.2& 30.6& 26.4&  27.8&58.3&27.8\\
&  add def.&30.6&  37.5& 26.4&  43.1&59.7&31.9\\
&  1-shot IcL&18.1&  26.4& 29.2&  31.9&48.6& 34.7\\
&  3-shot IcL&22.2&  30.6& 27.8&  34.7&65.3& 48.6\\

\hline
\multirow{4}{*}{Maximal root set}&  Basic&25.0& 24.3& 38.2&  34.7&25.7&26.4\\
&  add def.&29.9&  22.2& 29.9&  27.1&43.1&27.1\\
&  1-shot IcL&29.2&  18.7& 38.9&  32.6&30.6& 31.9\\
&  3-shot IcL&31.9&  26.4& 45.1&  34.7&34.0& 40.3\\
\hline
\multicolumn{8}{c}{\cellcolor{orange!10}\emph{Advanced Task}}\\
\hline
\multirow{4}{*}{Frontdoor adjustment set}&  Basic&31.9& 29.2& 34.7&  31.9&62.5&45.8\\
&  add def.&40.3&  38.9& 41.7&  47.2&65.3&43.1\\
&  1-shot IcL&45.8&  33.3& 44.4&  44.4&69.4& 45.8\\
&  3-shot IcL&44.4&  30.6& 48.6&  45.8&63.9& 51.4\\

\hline
\end{tabular}
\end{table*}
\end{center}

\end{document}